\documentclass[letterpaper, 10 pt, conference]{ieeeconf}  
\IEEEoverridecommandlockouts   
\usepackage{times}
\usepackage{epsfig}
\usepackage{graphicx}
\usepackage{amssymb}
\usepackage{amsmath}
\usepackage{amsthm}

\usepackage{xcolor}
\usepackage{gensymb}
\usepackage[hyphens]{url}
\usepackage{multirow}
\usepackage{multicol}
\usepackage{tabularx}
\usepackage{dsfont}
\usepackage{url}
\usepackage{color}
\usepackage{subcaption}
\usepackage{caption}
\usepackage{booktabs}
\usepackage{dcolumn}
\usepackage{colortbl}
\usepackage{pifont}
\usepackage{algorithm}
\usepackage{algpseudocode}

\usepackage[pagebackref=true,breaklinks=true,letterpaper=true,colorlinks,bookmarks=false]{hyperref}

\newcommand{\cmark}{\ding{51}}%

\colorlet{lightgray}{gray!20}

\begin{document}

\title{HM-ViT: Hetero-modal Vehicle-to-Vehicle Cooperative perception with vision transformer}

\author{Hao Xiang$^{1}$, Runsheng Xu$^{1}$, Jiaqi Ma$^{1\dagger}$
\thanks{$^{1}$ University of California, Los Angeles, CA, USA.}
\thanks{$\dagger$ Corresponding author: \texttt{jiaqima@ucla.edu}.}
}

\maketitle

\begin{abstract}
   Vehicle-to-Vehicle technologies have enabled autonomous vehicles to share information to see through occlusions, greatly enhancing perception performance. Nevertheless, existing works all focused on homogeneous traffic where vehicles are equipped with the same type of sensors, which significantly hampers the scale of collaboration and benefit of cross-modality interactions. In this paper, we investigate the multi-agent hetero-modal cooperative perception problem where agents may have distinct sensor modalities. We present HM-ViT, the first unified multi-agent hetero-modal cooperative perception framework that can collaboratively predict 3D objects for highly dynamic vehicle-to-vehicle (V2V) collaborations with varying numbers and types of agents. To effectively fuse features from multi-view images and LiDAR point clouds, we design a novel heterogeneous 3D graph transformer to jointly reason inter-agent and intra-agent interactions. The extensive experiments on the V2V perception dataset OPV2V demonstrate that the HM-ViT outperforms SOTA cooperative perception methods for V2V hetero-modal cooperative perception. We will release codes to facilitate future research.
\end{abstract}

\section{Introduction}
Recent advances in Vehicle-to-Vehicle (V2V) communication technology and intelligent transportation systems~\cite{guo2020leveraging,raboy2021proof, ma2020empirical, guo2020evaluating, chen2022prediction, kitajima2022nationwide, papathanasopoulou2022data} have allowed autonomous vehicles (AVs) to share sensory information, enabling them to perceive their surroundings better~\cite{betz2022autonomous,shladover2021opportunities}. With the rapid growth of autonomous driving, V2V perception systems have the potential to be deployed at scale and create a safer transportation system. Cooperative perception systems, as shown in recent studies~\cite{xu2022opv2v,xu2022v2x,wang2020v2vnet}, can intelligently aggregate features from multiple vehicles within the communication range to enhance visual reasoning and overall performance.

\begin{figure}
    \centering
    \vspace{1em}
    \includegraphics[width=3.3in]{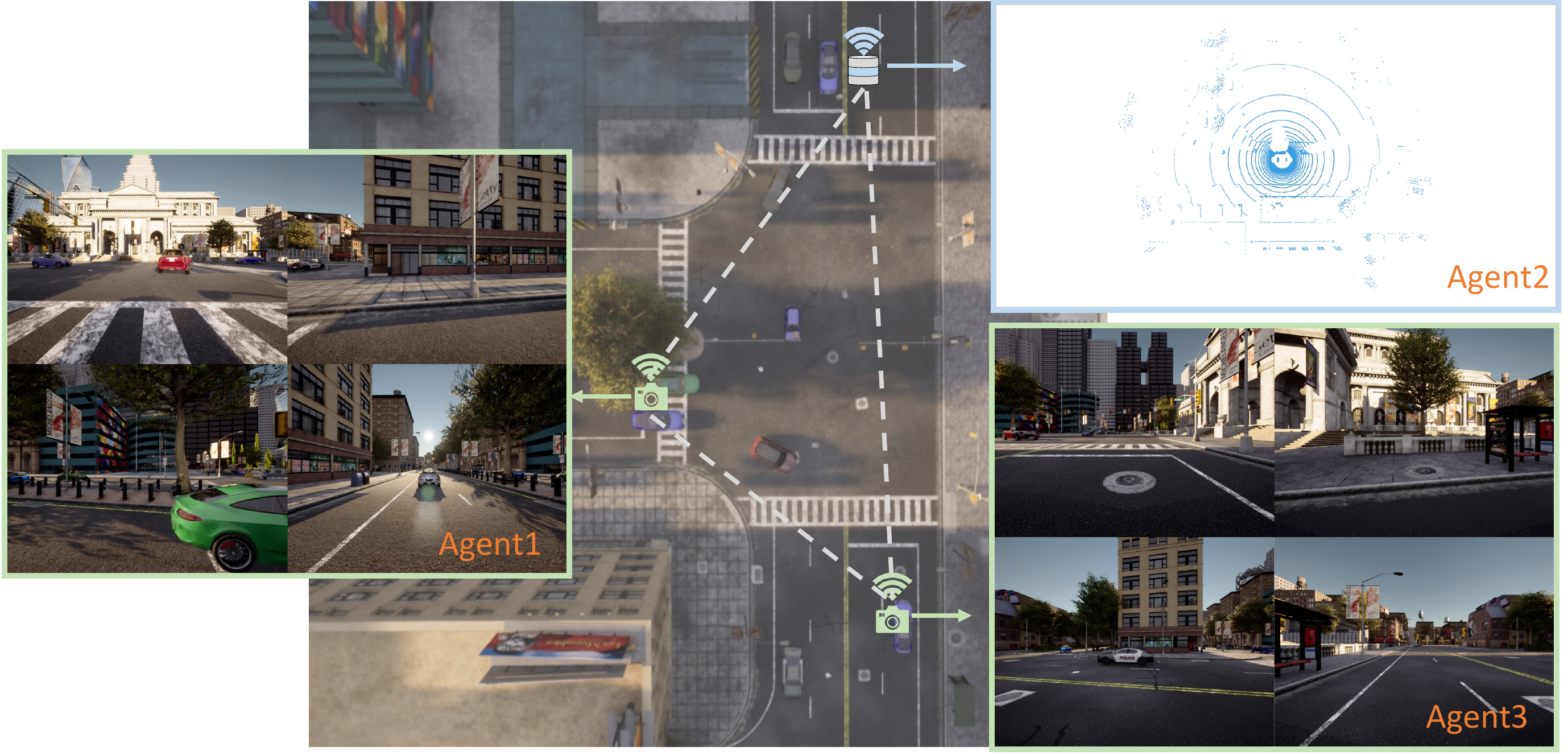}
    \caption{Illustration of multi-agent hetero-modal V2V systems where each agent may be equipped with either LiDAR or multi-view cameras.}
    \label{fig:sensor_config}
    \vspace{-1em}
\end{figure}

Despite the rapid growth in this field, previous studies~\cite{xu2022opv2v, wang2020v2vnet, xu2022cobevt, li2021learning, valiente2022robustness, xu2023v2v4real} have primarily focused on homogeneous multi-agent cooperative perception, where all agents are equipped with the same type of sensors. In reality, however, agents may have different sensor modalities (hetero-modality) due to the cost and sensor preferences of ADS developers and car makers. 
As shown in Fig.~\ref{fig:sensor_config}, some agents are equipped with only LiDARs (LiDAR agents), while others only have multiple cameras (camera agents). Enabling collaboration between these heterogeneous agents could improve the sensing capability by allowing agents to see through occlusions and increase the scale and reliability of V2V systems. Additionally, LiDAR agents can provide accurate geometric information, while camera agents can provide rich semantic context. Thus, the collaboration between these agents could leverage the distinct but complementary environment attributes captured by each sensor modality to enhance the V2V perception systems. Furthermore, compared with the single-agent solution where multiple LiDARs and cameras are installed in a single vehicle, distributing different types of sensors across distinct agents could also potentially decrease the costs for each agent while still achieving satisfying performance.  Nevertheless, whether, when, and how multi-agent hetero-modal V2V cooperation can benefit the perception system of heterogeneous traffic has not yet been studied. 

\begin{figure}
    \centering
    \includegraphics[width=3.2in]{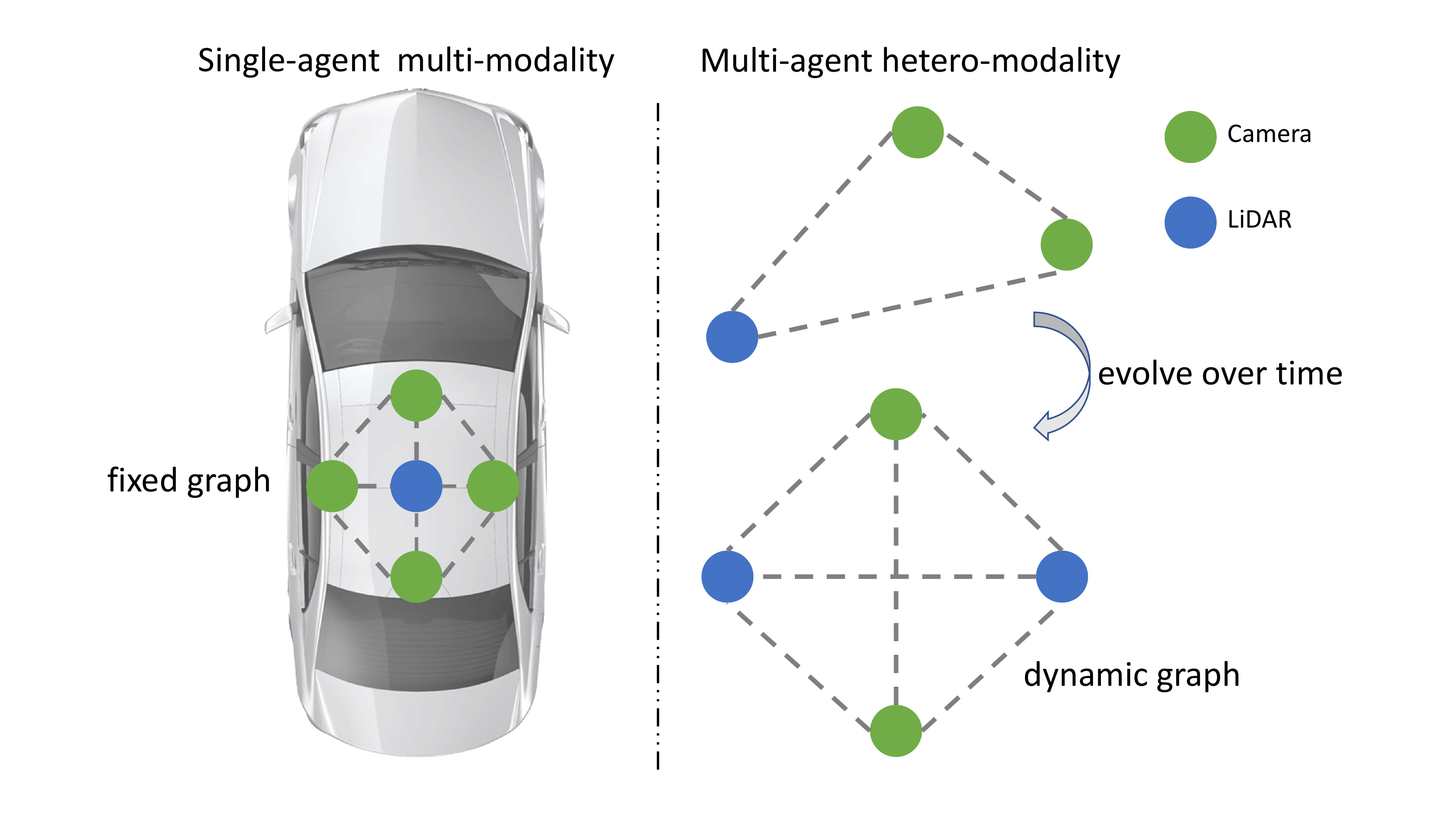}
    \caption{Comparison of the single-agent multi-modal system and multi-agent hetero-modal system. The graph structure of the former is fixed whereas for the latter the graph is both dynamic and heterogeneous.}
    \label{fig:comparison}
\end{figure}

In this work, we address the multi-agent hetero-modal cooperative perception problem where each agent could have distinct sensor types and share/receive information with each other. Notably, as shown in Fig.~\ref{fig:comparison}, this multi-agent hetero-modal setting is distinct from the single-agent multi-modal setting. In the hetero-modal setting, the agent sensors form a dynamic heterogeneous graph where the existence and types of sensors are random, and the relative poses vary from scene to scene. In contrast, the sensor types/numbers and relative positions (extrinsics) between sensors are fixed in the single-agent multi-modal setting. Existing multi-modal methods heavily rely on these assumptions, and most of existing works~\cite{vora2020pointpainting,wang2021pointaugmenting,bai2022transfusion,zhao2021lif,xu2022fusionrcnn} transform LiDAR points or 3D proposals onto the image plane to index 2D features. Their network architectures build upon the co-existence of both LiDAR and camera inputs with fixed geometric relationships. However, the dynamic nature of hetero-modal V2V perception requires a flexible architecture that can handle varying agent numbers and types, and the transmitted neural features are also spatially misaligned. Moreover, there are semantic discrepancies in the transmitted features between camera agents and LiDAR agents. Hence, these unique characteristics pose significant challenges for designing the multi-agent hetero-modal cooperative system and prevent adapting existing multi-modal fusion methods to this new problem.

\begin{figure*}[!t]
\centering
    \centering{\includegraphics[width=1\linewidth]{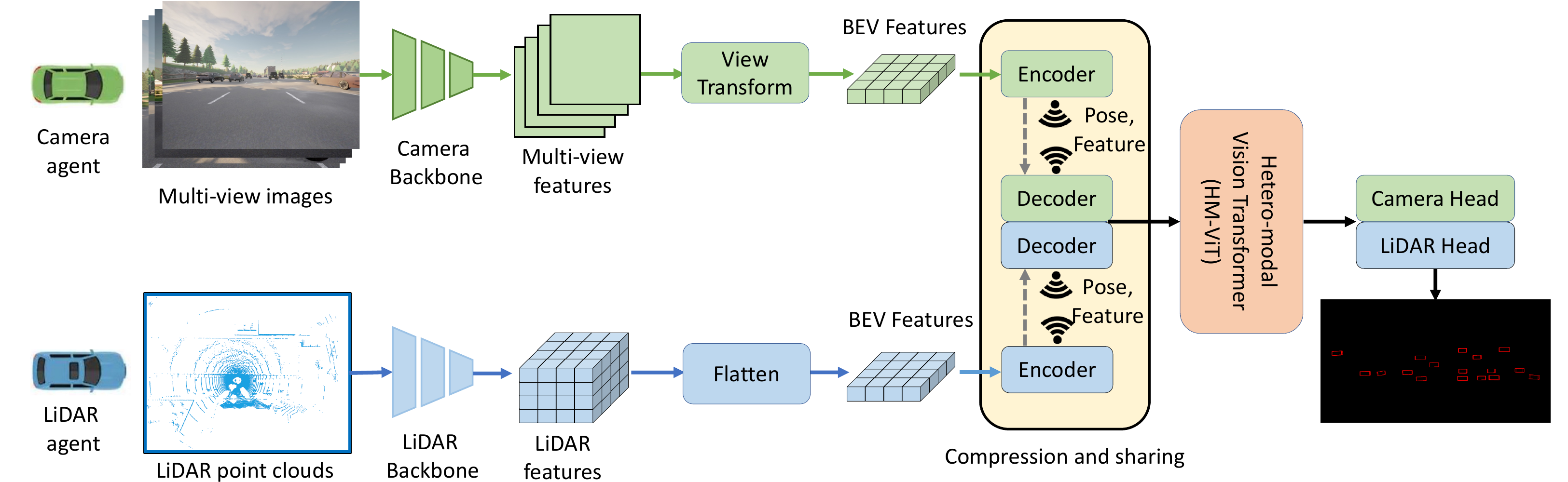}}
\caption{Overview of proposed hetero-modal V2V cooperative perception system. Each agent first produces BEV features through modality-specific feature extractors~(Sec.~\ref{sec:feature_extraction}). The BEV features are then compressed and shared~(Sec.~\ref{sec:hm-vit}) with neighboring connected agents and the received features are decompressed in the ego agent side and fed into hetero-modal vision transformer to conduct graph-structured feature fusion~(Sec.~\ref{sec:h3gat} and Sec.~\ref{sec:hm-vit}). The refined features are finally passed into the hetero-modal detection head to predict 3D bounding boxes (Sec.~\ref{sec:head}).  }
\label{fig:overview}
\end{figure*}

To enable collaboration between heterogeneous agents in V2V systems, we propose HM-ViT, the first unified cooperative perception framework that can leverage and fuse distributed information for hetero-modal V2V perception via a spatial-aware 3D heterogeneous vision transformer. Fig.~\ref{fig:overview} demonstrates the overall framework. Each agent first generates bird's eye view (BEV) representations through modality-specific encoders and then shares compressed features with neighboring agents. Afterward, the received features are decompressed and aggregated via the proposed HM-ViT, which conducts joint local and global heterogeneous 3D attentions with the consideration of both node and edge types. Our extensive experiments show that the HM-ViT can significantly improve the perception capability of camera agents and LiDAR agents over the single-agent baseline and outperforms SOTA cooperative perception methods by a large margin.  In particular, for camera agents, performance can be boosted from 2.1\% to 53.2\% at AP@0.7 with the collaboration of LiDAR agents, a \textbf{23-fold} improvement. Our primary contributions can be summarized as follows:
\begin{itemize}
    \item We present the novel transformer framework (HM-ViT) for multi-agent hetero-modal cooperative perception, capable of capturing the modality-specific characteristics and heterogeneous 3D interactions. The proposed model exhibits superior flexibility and robustness with state-of-the-art performance on highly dynamic heterogeneous traffic involving varying agent numbers/types.
    \item We propose a generic heterogeneous 3D graph attention (H$^3$GAT), tailored for extracting inter-agent and intra-agent heterogeneous interactions. We instantiate two such attentions -- local attention (H$^3$GAT-L) and global attention (H$^3$GAT-G) for capturing both local and global visual cues. 
    \item We conduct extensive benchmark experiments by varying sensor modalities, demonstrating the strong performance of the proposed method for hetero-modal V2V perception tasks. We will release all the codes and baselines to facilitate future research. 
 \end{itemize}


\section{Related work}
\noindent\textbf{V2V perception.} V2V perception aims to enhance the perception performance of autonomous vehicles by leveraging shared information from other connected vehicles. Existing works have primarily focused on LiDAR-based 3D object perception. The pioneer cooperative perception methods transmit raw sensing observation (i.e., early fusion) or perception outputs (i.e., late fusion), whereas recent works~\cite{xu2022opv2v, wang2020v2vnet,hu2022where2comm,xiang2022v2xp,lei2022latency,xu2022v2x} are exploring the use of circulating intermediate neural features for achieving better performance-bandwidth trade-off.  V2VNet~\cite{wang2020v2vnet} employs graph neural networks to aggregate the shared neural features for joint detection and prediction. AttFuse~\cite{xu2022opv2v} uses single-head attention to model the per-location multi-agent interaction. Disconet~\cite{li2021learning} presents a matrix-valued edge weight for learning the interactions and a teacher-student learning framework to facilitate the training. V2X-ViT explores~\cite{xu2022v2x} vision transformer for vehicle-to-everything cooperation via window attention and heterogeneous self-attention. CoBEVT~\cite{xu2022cobevt} presents a generic transformer framework for camera-based BEV semantic segmentation.

\noindent However, none of the existing works explored multi-agent multi-camera 3D object detection, let alone multi-agent hetero-modal perception.
In contrast to existing methods, HM-ViT is the first to employ sparse heterogeneous local and global attentions to capture the 3D inter-agent and intra-agent interactions in a computationally efficient manner.   


\noindent\textbf{Camera-based 3D object detection.} Early works~\cite{brazil2019m3d,chen2016monocular,simonelli2019disentangling} mainly focus on monocular 3D detection but a single camera can only provide a 2D view of the scene, and inferring 3D from 2D is intrinsically hard. 
Recent development of self-driving datasets~\cite{caesar2020nuscenes,sun2020scalability,chang2019argoverse} featured with full sensor suits has enabled the research direction of 3D object detection from multiple cameras. DETR3D~\cite{wang2022detr3d} proposes a 3D-2D query paradigm for extracting 3D features from 2D multi-view images. Graph-DETR3D~\cite{chen2022graph} further utilizes graph structure learning to enhance the representation at the border regions. Alternatively, another stream of works~\cite{huang2021bevdet, philion2020lift,xie2022m,li2022bevformer} focuses on aggregating BEV features from multi-view cameras for conducting downstream perception tasks. LSS~\cite{philion2020lift} lifts the 2D features to 3D frustum via latent depth and then splats frustums into a BEV grid.  M$^2$BEV~\cite{xie2022m} further extends LSS with less memory consumption and conducts detection and segmentation. BEVFormer~\cite{li2022bevformer} constructs BEV queries and explores spatial cross-attention and temporal self-attention to recurrently refine the BEV features, achieving SOTA performance on both NuScenes~\cite{caesar2020nuscenes} and Waymo Open Datasets~\cite{sun2020scalability}.

\begin{figure*}[!t]
\centering
\centering{\includegraphics[width=1\linewidth]{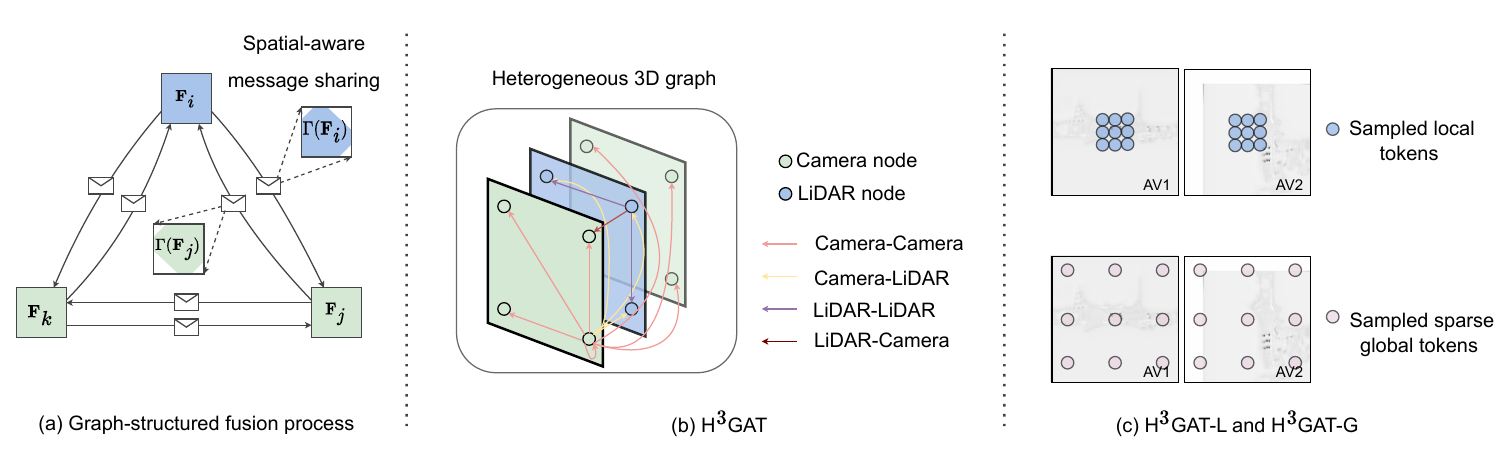}}
\caption{HM-ViT architecture. (a) Graph-structured fusion process. (b) Heterogeneous 3D graph attention (H$^3$GAT). (c) Illustration of sampled tokens for H$^3$GAT-L and H$^3$GAT-G.}
\label{fig:hm-vit}
\end{figure*}
\noindent\textbf{Multi-modal fusion.}
Existing multi-sensor fusion methods can be divided into point-level fusion, proposal-level fusion, and BEV-level fusion. Point-level fusion decorates the input from one modality with attributes from the other modality. PointPainting~\cite{vora2020pointpainting} and PaintAugmenting~\cite{wang2021pointaugmenting} decorate the LiDAR point clouds with semantic segmentation scores and 2D CNN image features respectively while \cite{deng2017amodal, alexandre20163d} project LiDAR onto the image plane to augment the RGB values with depth information and conduct detection. On the other hand, proposal-level fusion generates proposals from one modality and then indexes features from the other modality for further refinement. MV3D~\cite{chen2017multi} produces object queries from LiDAR BEV and then extracts the features from camera data and LiDAR front view.~\cite{qi2018frustum, wang2019frustum} lift 2D bounding boxes to frustum and conduct 3D object detection from the frustum of point clouds. Conversely, BEV-level fusion converts features from different modalities to unified BEV representations, preserving both geometric and semantic information. BEVFusions~\cite{liu2022bevfusion,liang2022bevfusion} concatenate BEV features from camera and LiDAR and fuse it via a fusion module.  
Hence, existing single-agent multi-modal fusion methods rely on the co-existence of both camera and LiDAR with fixed geometric relationships, which is unsuitable for our multi-agent hetero-modal cooperative perception problem with a dynamic heterogeneous collaboration graph. 

\section{Methodology}
In this paper, we explore the multi-agent hetero-modal cooperative perception, where each AV is equipped with either a LiDAR or multiple cameras. Our goal is to create a robust and flexible cooperative perception system that allows for efficient collaboration between any number of agents with varying sensor types, ultimately improving the perception capabilities of the vehicle in a unified end-to-end fashion. The pipeline, illustrated in \ref{fig:overview},  includes modality-specific feature extraction, compression and sharing, HM-ViT for feature fusion, and hetero-modal detection head. 

\subsection{Modality-specific feature extraction}
\label{sec:feature_extraction}
\noindent\textbf{LiDAR stem: }We leverage PointPillar~\cite{lang2019pointpillars} to process point clouds for each LiDAR agent. The raw point cloud is converted to a 2D pseudo-image, flattened along the height dimension, and fed into 2D convolutional neural networks to produce the salient feature map $\mathbf{F}_j\in\mathcal{R}^{H\times W\times C}$, which is compressed and shared with all the neighboring agents. 

\noindent\textbf{Camera stem: }Each camera agent is equipped with $m$ monocular cameras. The sensing observation of $i$-th agent includes the input images $I_k^i\in\mathcal{R}^{h\times w \times 3}$ and the known projection matrix $P_k^i\in\mathcal{R}^{3\times 4}$ that maps 3D reference points to different image views. Our goal is to generate a BEV feature representation $\mathbf{F}_i\in\mathcal{R}^{H\times W\times C}$ that is amenable for feature fusion with other collaborators. In this work, we adopt similar architecture to BEVFormer~\cite{li2022bevformer} with no temporal information for feature extraction. For a faster running time, we adopt ResNet50 to extract 2D image features and then adopt a learnable 2D BEV query to inquire spatial information from the encoded multi-view features via spatial cross attention and projection matrices. The resulting refined BEV feature $\mathbf{F}_i$ is centered around agent $i$ and shared with connected AVs.

\subsection{Heterogeneous 3D Graph Attention (H$^{3}$GAT)}
\label{sec:h3gat}
To account for the distinct characteristics of BEV features extracted from different sensor modalities, the learning process of each modality must be distinguished, and the cross-modality interactions between multiple agents should vary.
To capture this heterogeneity, we present a novel heterogeneous 3D graph attention ({H$^{3}$GAT}), in which nodes and edges are type-dependent to reason spatial interactions and cross-agent relations jointly. We encode both local and global interactions to better capture the 3D ambiguity in BEV feature space. Local attention can help preserve object details, while global attention can provide a better understanding of environmental contexts such as road topology and traffic density. 

As shown in Fig.~\ref{fig:hm-vit}b, We build a 3D heterogeneous collaboration graph.  Each node ${v}\left(i,x\right)=\mathbf{F}^{i}_{x}\in\mathcal{R}^{C}$ is a feature vector of agent $i$'s feature map at spatial location $x\in\mathcal{R}^2$.  3D heterogeneous graph attention is performed for spatially connected nodes in the BEV feature space.  Depending on the definition of spatial connectivity, we will derive local attention and global attention. Here for notation simplicity, we only derive single-head equations but in real implementations, multi-head variants are used. Formally, we first project feature vectors onto different feature spaces to form query, key, and value vectors: 
\begin{align}
    \mathbf{Q}^{j}_x &= \text{Dense}_{\tau_j}\mathbf{F}^{j}_{x}\\
    \mathbf{K}^{j}_x &= \text{Dense}_{\tau_j}\mathbf{F}^{j}_{x}\\
    \mathbf{V}^{j}_x &= \text{Dense}_{e_{ij}}\mathbf{F}^{j}_{x}
\end{align}
where the $\text{Dense}_{\qedsymbol{}}$ is a set of linear layers indexed by subscript $\qedsymbol$. 
For the query and key vectors, we use linear projectors $\text{Dense}_{\tau_j}$ indexed by node type $\tau_j$ to extract modality-specific features. For the value vector, we index the projector via edge type $\text{Dense}_{e_{ij}}$ to reflect the heterogeneity of cross-modality multi-agent interactions. Denote the set of connected nodes of $v\left(i,x\right)$ as $\mathcal{N}\left(i, x\right)$. Then the attention is operated as follows:
\begin{align}
    \mathbf{a}\left(j, y\right)&=\underset{(j,y)\in \mathcal{N}\left(i, x\right)}{\text{Softmax}}\left(\mathbf{Q}^{i}_x\mathbf{W}_{e_{ij}}\mathbf{K}^{j}_{y}\right)\\
    \mathbf{F}^i_x&=\sum_{(j,y)\in \mathcal{N}\left(i, x\right)}\mathbf{a}\left(j,y\right)\mathbf{V}^{j}_y
\end{align}
where $\mathbf{W}_{e_{ij}}\in\mathcal{R}^{C\times C}$ is used to adjust the dot product of $\mathbf{Q}^{i}_x$ and $\mathbf{K}^{j}_y$ to further encode the heterogeneity of edges. 
\begin{figure}
    \centering
    \includegraphics[width=3.2in]{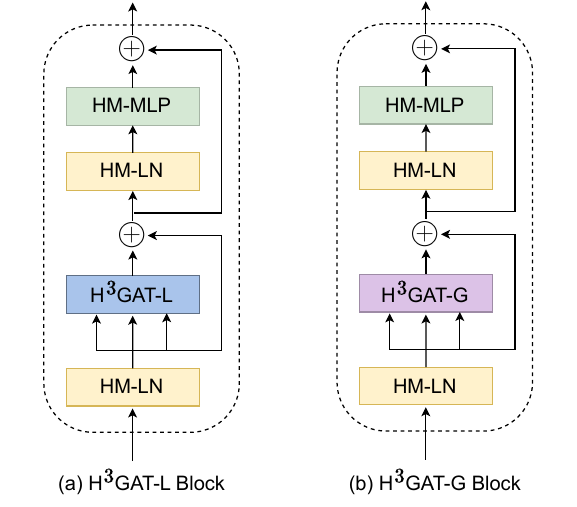}
    \caption{Transformer blocks for local and global attentions.}
    \label{fig:block}
    \vspace{-1em}
\end{figure}

Depending on how the nodes are sampled (Fig.~\ref{fig:hm-vit}c), we design two types of attentions: local attention (\textbf{H${^3}$GAT-L}) which performs local window-based attention and global attention (\textbf{H${^3}$GAT-G}) which performs sparse global grid-based attention. The local interactions can help preserve spatial cues and provide reliable estimates while the global reasoning can help understand global semantic context.

Both H${^3}$GAT-L and H${^3}$GAT-G can be implemented efficiently by decomposing the spatial axes. More specifically, we stack all the agents' features to $\mathbf{F}\in\mathcal{R}^{N\times H\times W\times C}$ where $N$ is the number of agents. For H${^3}$GAT-L, we decompose the feature map into 3D non-overlapping windows along the first axis~\cite{tu2022maxim,tu2022maxvit}, each of size $N\times P\times P$. The partitioned tensor has the shape $(\frac{H}{P}\times \frac{W}{P},N\times P^2, C)$ where the heterogeneous 3D local graph attention is conducted for $NP^2$ tokens within the same window. Similarly, for H${^3}$GAT-G , we swap the axis and partition the tensor into the shape $({N\times P^2},\frac{H}{P}\times\frac{W}{P},C)$ where the attention is operated for these sparsely sampled $\frac{H}{P}\times\frac{W}{P}$ grids, which can capture sparse global information. 

To integrate this local and global attention into transformer architecture, we further present a heterogeneous normalization layer (\textbf{HM-LN}) and heterogeneous MLP (\textbf{HM-MLP}) which use type-dependent parameters. As shown in Fig.~\ref{fig:block}, we first pass all the features into the HM-LN where different statistics are calculated and used as per each agent's modality type. Afterward, we feed the normalized features into a heterogeneous 3D graph attention (H$^{3}$GAT-L/H$^{3}$GAT-G) to jointly reason heterogeneous inter-agent and intra-agent interactions. Then, we pass the fused features to another HM-LN followed by a Hetero-modal MLP layer where different sets of parameters are used for camera and LiDAR features.  By carefully designing these components, we can maintain modality-specific characteristics throughout the fusion process while benefiting from cross-modality multi-agent interactions. 
\begin{figure}
\vspace{-2mm}
\begin{algorithm}[H]
    \caption{\textbf{Multi-agent hetero-modal fusion process}}\label{alg:fusion}
    \begin{algorithmic}[1]
    \State \textbf{Input: }decompressed feature $\mathbf{F}_i$, pose $x_{i}$ for each agent
    \State $\mathbf{F}_i^{(0)} = \mathbf{F}_i$
        \For{$l=1...\dots L$}
        \For{each agent $i$}\Comment{Process in parallel}
            \State $\mathbf{F}_{j\rightarrow i}^{(l-1)}={\Gamma}_{j\rightarrow i}\left(\mathbf{F}_j^{(l-1)}\right)$\Comment{Spatially transform neighboring agents' features}
            \State $\mathbf{F}_i^{(l)}=\text{H${}^3$GAT-L Block}(\{\mathbf{F}_{j\rightarrow i}^{(l-1)}\})$ \Comment{Update node via local attention}
        \EndFor
        \For{each agent $i$}\Comment{Process in parallel}
            \State $\mathbf{F}_{j\rightarrow i}^{(l)}={\Gamma}_{j\rightarrow i}\left(\mathbf{F}_j^{(l)}\right)$\Comment{Spatially transform neighboring agents' features}
            \State $\mathbf{F}_i^{(l)}=\text{H${}^3$GAT-G Block}(\{\mathbf{F}_{j\rightarrow i}^{(l)}\})$ \Comment{Update node via global attention}
        \EndFor
        \EndFor
        \State $\mathbf{F}_i$ = HM-MLP$\left(\mathbf{F}_i^{(L)}\right)$\Comment{Output updated features}
    \end{algorithmic} 
\end{algorithm}
\vspace{-4mm}
\end{figure}
\subsection{Hetero-modal Vision Transformer}
\label{sec:hm-vit}
\noindent\textbf{Compression and sharing:} To reduce the transmission bandwidth, a series of $1\times1$ convolutions is applied to reduce the transmitted feature size along the channel dimension. Together with the intermediate features, each agent's pose $x_i$ is also circulated within the collaboration graph. The ego agent will receive these features and decompress them back to the original size via another convolutional network. For processing intermediate features of the camera agent and LiDAR agent, we leverage distinct parameters in the compression and decompression modules to preserve the modality-specific characteristics. 

\begin{table*}[!t]
\centering
\setlength{\tabcolsep}{8pt}
\renewcommand{\arraystretch}{1.0}
\label{table:benchmark}
\begin{tabular}{l|cc|cc|cc}
\cellcolor{lightgray} & \multicolumn{2}{c|}{\cellcolor{lightgray} V2V-C}&\multicolumn{2}{c|}{\cellcolor{lightgray} V2V-L}&\multicolumn{2}{c}{\cellcolor{lightgray} V2V-H}\\ 
\cline{2-3}\cline{4-5}\cline{6-7}
{\cellcolor{lightgray} Models}
& \cellcolor{lightgray} AP@0.5& \cellcolor{lightgray} AP@0.7& \cellcolor{lightgray} AP@0.5& \cellcolor{lightgray} AP@0.7& \cellcolor{lightgray} AP@0.5& \cellcolor{lightgray} AP@0.7\\ \toprule
No Fusion&0.094&0.021&0.524&0.363&0.284&0.157\\
Late Fusion&0.231&0.070&0.770&0.606&0.502&0.308\\
\bottomrule
V2VNet~\cite{wang2020v2vnet}&0.329&0.125&0.820&0.645&0.650&0.366\\
DiscoNet~\cite{li2021learning}&0.287&0.115&0.741&0.590&0.624&0.385\\
AttFuse~\cite{xu2022opv2v}&0.261&0.095&0.801&0.644&0.647&0.390\\
CoBEVT~\cite{xu2022cobevt}&0.317&0.122&0.828&0.637&0.674&0.416\\
V2X-ViT~\cite{xu2022v2x}&0.332&0.125&0.833&0.679&0.671&0.427\\
\bottomrule
HM-ViT&\textbf{0.355}&\textbf{0.142}&\textbf{0.853}&\textbf{0.763}&\textbf{0.695}&\textbf{0.515}\\
\bottomrule
\end{tabular}
\caption{Evaluation of V2V perception methods on OPV2V dataset. }
\label{table:benchmark}
\end{table*}

\noindent\textbf{Graph-structured feature fusion: }The received BEV features are centered around different spatial locations as each agent perceives the dynamic environment from different view points. To this end, we present a graph-structured fusion process (Fig.~\ref{fig:hm-vit}a): each node maintains a state representation of an agent in its own coordinate frame, and for a fixed number of iterations, spatially warped messages are shared between nodes and the node states are updated based on the aggregated features via the transformer blocks. 

The overall process is summarized in Alg.~\ref{alg:fusion}. During each iteration, we have two cascaded node updates which capture the local and global heterogeneous interactions respectively. For each node, we first spatially transform ~\cite{jaderberg2015spatial} neighboring nodes' features to its center $\mathbf{F}_{j\rightarrow i}={\Gamma}_{j\rightarrow i}\left(\mathbf{F}_j\right)$. When the transmitting node is the receiving node itself, the transformation matrix is an identity matrix and thus $\mathbf{F}_{i\rightarrow i}=\mathbf{F}_i$. These spatial aligned feature maps $\mathbf{F}_{j\rightarrow i}$ are then shared with agent $i$ to update its state representation via the aggregation module. We adopt H$^{3}$GAT-L Block as our first aggregation module to capture the local heterogeneous interactions and leverage H$^{3}$GAT-G Block for the second module to further refine the states with global cues. Within each transformer block, we also adopt a mask to mask out non-overlapping areas between the field of views when computing the attention scores. Note that each agent's state update can be processed in parallel for better efficiency. After L such iterations, we pass the features to a hetero-modal MLP to further refine the feature representation. Throughout the whole fusion process, the modality-specific statistics are maintained. 




\begin{figure*}[!t]
\centering
\includegraphics[width=0.98\textwidth]{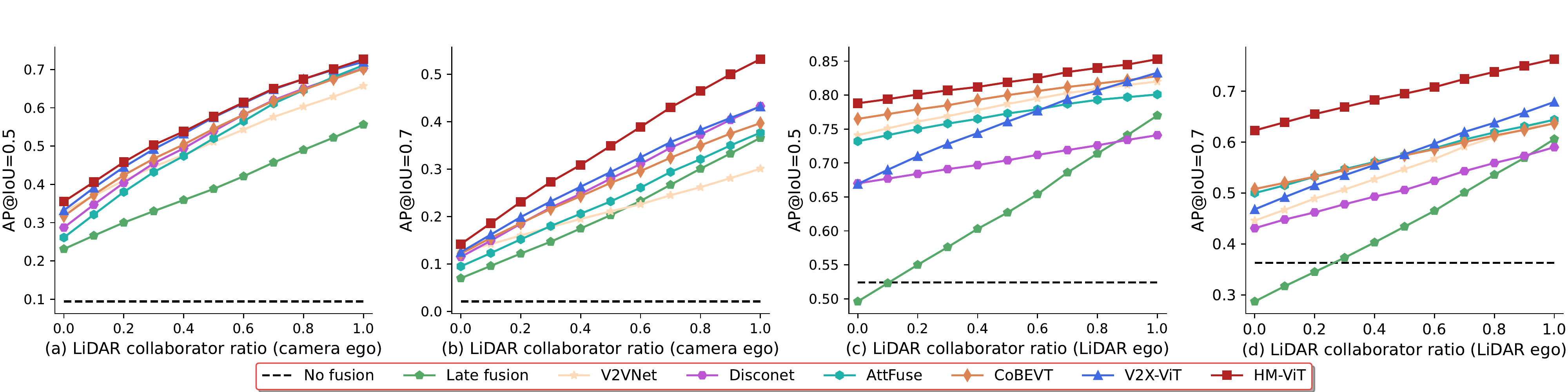}
\caption{Agent modality ratio experiment. The x-axis is the ratio of LiDAR collaborators among all the collaborators. In (a) and (b), ego vehicles are equipped with cameras. In (c) and (d), ego vehicles are equipped with LiDARs. }
\label{fig:ratio}
\vspace{-2mm}
\end{figure*}
\subsection{Hetero-modal Head}
\label{sec:head}
As camera and LiDAR contain distinct characteristics, we design a hetero-modal head where a different set of parameters are applied for camera and LiDAR ego vehicles to generate the final predictions.  More specifically, the final fused feature maps are passed to a series of $3\times 3$ convolutions with batch normalization and ReLU for feature refinement. Then, we adopt a 1×1 convolution layer to generate the regression and classification predictions. Smooth $\ell_1$ loss is utilized for regression and a focal loss~\cite{lin2017focal} is used for classification.

\section{Experiments}
\subsection{Datasets and Evaluation}
\noindent\textbf{OPV2V. } OP2V~\cite{xu2022opv2v} is a large-scale multi-modal cooperative V2V perception dataset collected in CARLA~\cite{dosovitskiy2017carla} and OpenCDA~\cite{xu2021opencda}. It contains over 70 driving scenarios of around 25 seconds duration each. Each scenario contains multiple connected AVs (2 to 7) and each AV is equipped with 1 LiDAR and 4 monocular cameras covering 360$\degree$ horizontal field of view (FoV). In our hetero-modal cooperative perception setting, we only use one type of sensor modality for each AV, leading to two types of agents: vehicles only equipped with multiple cameras (camera agent), and vehicles only equipped with LiDARs (LiDAR agent).


\begin{figure}
    \centering
    \includegraphics[width=0.475\textwidth]{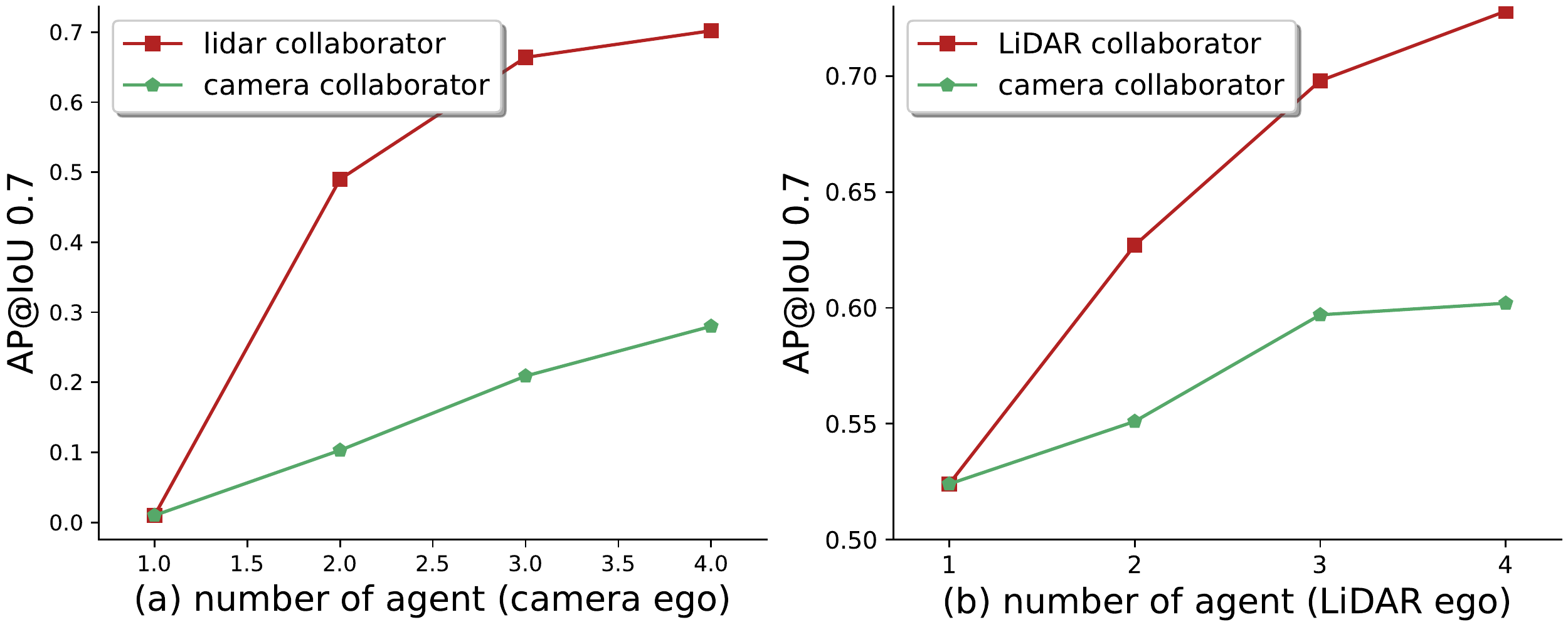}
    \caption{Ablation study on number of agent. The LiDAR/camera collaborator refers to the case where the collaborators are equipped with LiDAR and Camera respectively. (a) ego vehicle is equipped with camera. (b) ego vehicle is equipped with LiDAR.}
    \label{fig:number}
    \vspace{-1em}
\end{figure}

\noindent\textbf{Evaluations. }We adopt Average precision (AP) at Intersection-over-Union (IoU) 0.5 and 0.7 to measure the perception performance. As each scenario consists of multiple AVs, a fixed agent is selected as the ego agent and the evaluation is conducted in the range of 100 m $\times$ 100 m around it.  Following~\cite{xu2022opv2v,xu2022v2x}, the train/validation/test splits are 6764/1981/2719. We evaluate models mainly under three configurations: 1) {V2V Camera-based 3D detection (\textbf{V2V-C})}  where AVs are only equipped with 4 cameras with 360 horizontal FoV, 2) {V2V LiDAR-based detection (\textbf{V2V-L})} where all the agents only have LiDAR sensors, and 3) {V2V Hetero-modal detection (\textbf{V2V-H})} where half of the agents only has cameras while the other half only has LiDARs. To further assess models' capability with dynamic sensor configurations, we also assess models with fixed ego sensor modality and varying collaborator sensor types. 
\subsection{Experimental Setups}
\noindent\textbf{Implementation details. }We use the PointPillar~\cite{lang2019pointpillars} as 3D backbones and a modified BEVFormer~\cite{li2022bevformer} for our camera stem. For BEVFormer, we adopt its variant with no temporal information and ResNet50~\cite{he2016deep} as image backbones for better computation efficiency and use a smaller grid resolution (0.4~m) to preserve fine-trained spatial details. The intermediate BEV feature map has a dimension of $128\times128\times256$. We conduct two iterations of graph-structured feature fusion and employ a window size of 8 for both local and global attentions. We adopt AdamW with a decay rate of $10^{-2}$ and cosine annealing learning rate scheduler to optimize the model. \\
\noindent\textbf{Training strategy. }We find that the intermediate fusion models won't converge if directly trained end-to-end under V2V-H and the resulting models usually only exhibit good performance for either camera perception or LiDAR perception but hardly for both. Instead, we first train the model on single modality configurations (i.e., V2V-C and V2V-L) until convergence for 40 epochs and then fine-tune the models under V2V-H for 10 epochs with fixed parameters of modality-specific backbones on 4 RTX3090 GPUs. This training strategy can help models converge well with reliable performance. 

\noindent\textbf{Compared methods. }We regard No Fusion as the baseline method. We also evaluate the \textit{Late Fusion}, which transmits the detection proposals and leverages Non-maximum suppression to generate the final predictions. For the intermediate cooperation methods, we benchmark five approaches: V2VNet~\cite{wang2020v2vnet}, DiscoNet~\cite{li2021learning}, AttFuse~\cite{xu2022opv2v}, CoBEVT~\cite{xu2022cobevt}, and V2X-ViT~\cite{xu2022v2x}. For a fair comparison, hetero-modal head is used for all the models.

\begin{figure*}[!t]
\vspace{2mm}
\centering
    \begin{subfigure}[c]{0.38\textwidth}
        \centering{\includegraphics[width=1\textwidth]{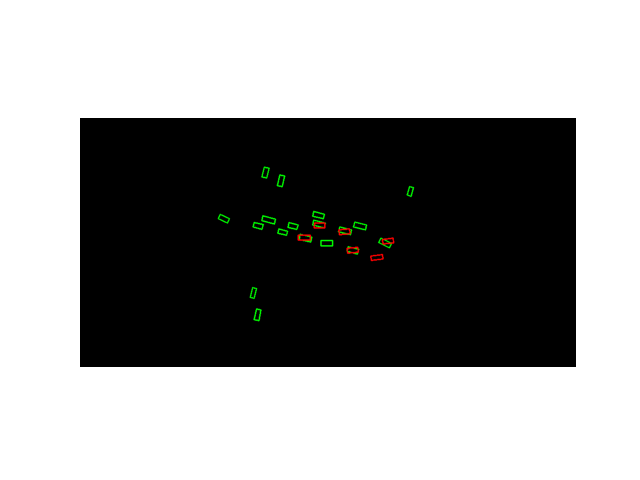}}
        \vspace{-3.5em}
        \caption{No Fusion (camera)}
        \label{fig:qualitive-a}
    \end{subfigure}\hspace{-3.7em}%
    \vspace{-3em}
    \begin{subfigure}[c]{0.38\textwidth}
        \centering{\includegraphics[width=1\textwidth]{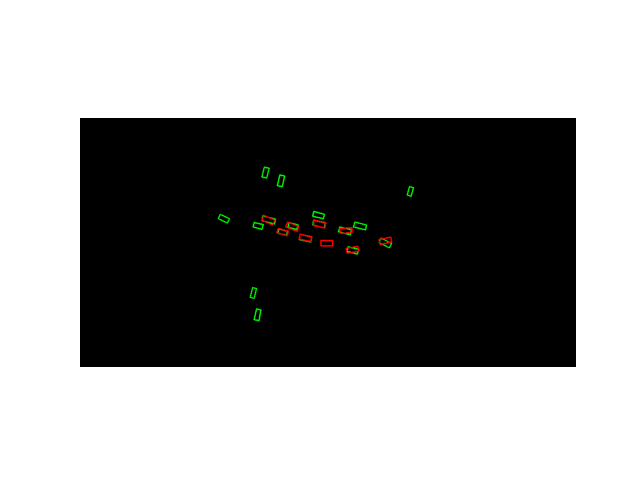}}
        \vspace{-3.5em}
        \caption{Pure camera fusion}
        \label{fig:qualitive-b}
    \end{subfigure}\hspace{-3.7em}%
    \begin{subfigure}[c]{0.38\textwidth}
        \centering{\includegraphics[width=1\textwidth]{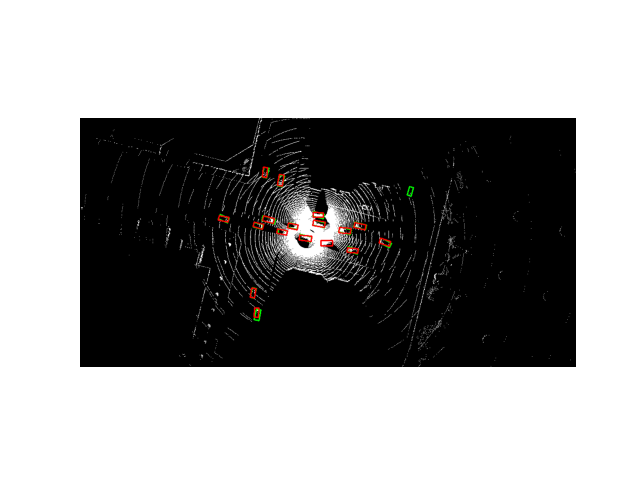}}
        \vspace{-3.5em}
        \caption{Camera ego with LiDAR collaborators}
        \label{fig:qualitive-c}
    \end{subfigure}\hspace{-3.7em}%
    \begin{subfigure}[c]{0.38\textwidth}
        \centering{\includegraphics[width=1\textwidth]{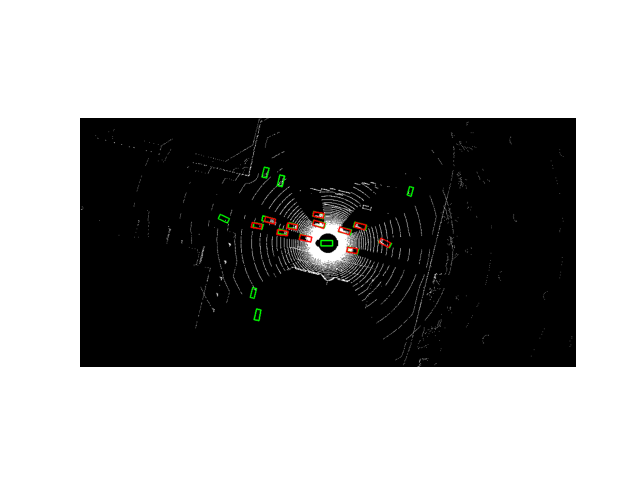}}
        \vspace{-4.8em}
        \caption{No Fusion (LiDAR)}
        \label{fig:qualitive-d}
    \end{subfigure}\hspace{-3.7em}%
    \begin{subfigure}[c]{0.38\textwidth}
        \centering{\includegraphics[width=1\textwidth]{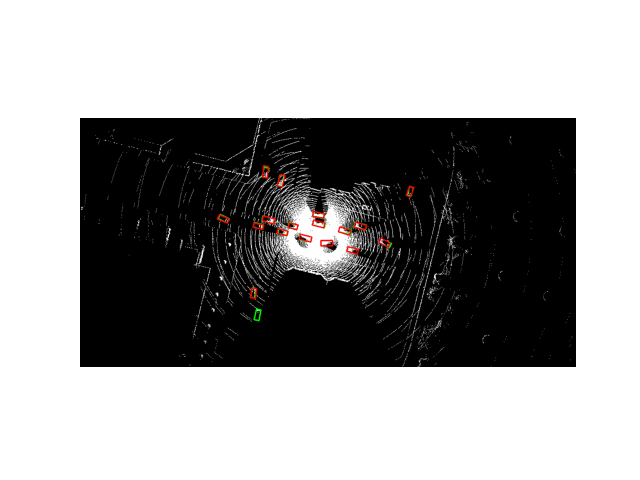}}
        \vspace{-4.8em}
        \caption{Pure LiDAR fusion}
        \label{fig:qualitive-e}
    \end{subfigure}\hspace{-3.7em}%
    \begin{subfigure}[c]{0.38\textwidth}
        \centering{\includegraphics[width=1\textwidth]{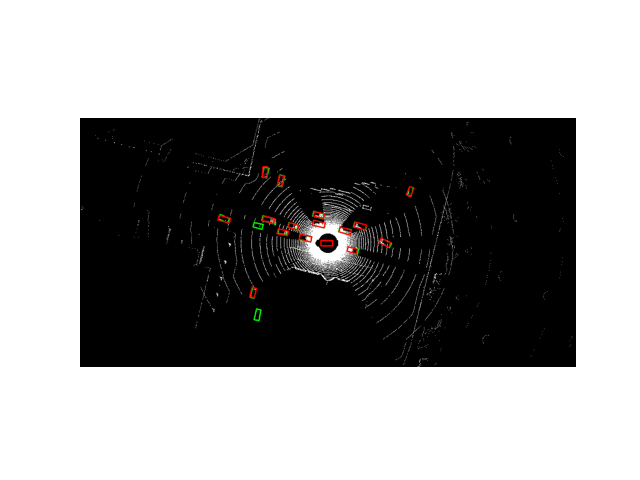}}
        \vspace{-4.8em}
        \caption{LiDAR ego with camera collaborators}
        \label{fig:qualitive-f}
    \end{subfigure}
    \caption{Qualitative visualizations for (a) No Fusion with camera ego vehicle, (b) pure camera-based V2V perception, (c) hetero-modal V2V perception with camera ego vehicle and LiDAR collaborators, (d) No Fusion with LiDAR ego vehicle, (e) pure LiDAR-based V2V perception, and (f) hetero-modal V2V perception with LiDAR ego vehicle and camera collaborators. The \textcolor{red}{red} and \textcolor{green}{green} boxes represent the  detection outputs and ground truth respectively. More visualizations can be found in the supplementary materials.}
    \label{fig:Qualitative}
    \vspace{-3mm}
\end{figure*}
\subsection{Quantitative evaluation}
\noindent\textbf{Main performance comparison. }Tab.~\ref{table:benchmark} demonstrates the performance comparisons on V2V-C, V2V-L and V2V-H. Under all three settings, all the cooperative methods outperform the No Fusion baseline and the intermediate fusion beats the classical Late Fusion, showing the great benefit of end-to-end V2V hetero-modal cooperative perception. The HM-ViT outperforms all the other SOTA intermediate fusion methods by at least $1.7\%$, $8.4\%$, $8.8\%$ in AP@0.7 under V2V-C, V2V-L, V2V-H settings respectively. 

\noindent\textbf{Agent modality ratio experiment. } As shown in Fig.~\ref{fig:ratio}, we fix the ego vehicle sensor modality and vary the ratio of collaborators' sensor modalities to evaluate the models' performance under various heterogeneous traffic scenarios. The larger LiDAR collaborator ratio corresponds to more vehicles only equipped with LiDARs while a smaller ratio means more vehicles only equipped with multiple cameras. The left two figures are the evaluation results for camera ego vehicles while the right two figures are the performance for LiDAR ego vehicles. Under most ratios, late fusion outperforms No Fusion however for LiDAR ego vehicle when most collaborators are camera agents, the Late Fusion performs poorer than No Fusion. We argue this is due to the fact that the camera predictions are usually noisy and merging proposals from different modalities equally could lead to ambiguity and thus deteriorate the performance. On the other hand, all the intermediate fusion methods outperform No Fusion by a large margin especially for the camera ego vehicles, demonstrating the great value of V2V cooperation between agents with different modalities.  Among all the compared methods, HM-ViT ranks first for both camera ego vehicles and LiDAR ego vehicles under all the ratios, illustrating the great capability of HM-ViT for capturing modality-specific characteristics and cross-modality multi-agent interactions. In contrast, other intermediate fusion methods only show good performance for a certain ratio range, which demonstrates the importance of heterogeneity for hetero-modal cooperative perception.
\begin{table}[]
    \centering
    \begin{tabular}{cccc}
\cellcolor{lightgray} {HM-MLP\&LN} &
 \cellcolor{lightgray} {H$^{3}$GAT-L} &
 \cellcolor{lightgray} {H$^{3}$GAT-G}&
\cellcolor{lightgray}AP@0.7 \\ \toprule
  &  &  & 0.404  \\
\cmark  &  & & 0.420  \\
\cmark & \cmark &  & 0.460 \\
 \cmark & \cmark & \cmark &  \textbf{0.515} \\
 \bottomrule
\end{tabular}
    \caption{Component Ablation study on the V2V-H setting}
    \label{tab:component}
    \vspace{-1em}
\end{table}

\noindent\textbf{Number of agent. }In this experiment, we investigate the effect of the number of agents on the perception performance of HM-ViT. As Fig.~\ref{fig:number} depicts, for both camera and LiDAR ego vehicles, the perception performance increases as more agents are involved in the cooperative perception and both LiDAR and camera collaborators can contribute to the performance gain for ego vehicles with different modalities, which again shows the benefit of hetero-modal V2V cooperation. Additionally, the increase rate generally decreases when increasing the number of agents and the LiDAR collaborators can bring more AP gains over the camera collaborators for both camera and LiDAR ego vehicles.  Notably, similar as Fig.~\ref{fig:ratio}a-b display, the camera ego vehicles' performance can be greatly improved with only a small number of LiDAR collaborators, demonstrating the great potential of reducing the cost for each vehicle when the V2V system is deployed at scale as we may only need to install expansive LiDARs for a small number of agents (e.g.,, infrastructure) while all the other agents only require relatively cheap camera sensors. 

\begin{table}[]
    \centering
    \begin{tabular}{c|c}
\cellcolor{lightgray}  Compression Rate&
 \cellcolor{lightgray} AP@0.7\\ \toprule
  0x& 0.515  \\
  8x&0.513\\
  16x& 0.470\\
  32x& 0.455\\ 
 \bottomrule
\end{tabular}
\caption{Compression rate effects for HM-ViT on V2V-H.}
\label{tab:compression}
\vspace{-1em}
\end{table}
\noindent\textbf{Component ablation study. }Here we investigate the key components of the proposed HM-ViT. As the layer normalization and MLP are usually combined together in typical transformer designs, thus we jointly evaluate the combined effect of HM-MLP and HM-LN (HM-MLP\&LN). As Tab.~\ref{tab:component} shows, all the proposed components improve the performance and local and global attentions can largely improve the AP@0.7 by 4\% and 5.5\%, proving the great benefit brought by jointly reasoning inter-agent and intra-agent heterogeneous interactions both locally and globally.

\noindent\textbf{Compression rate. }
Tab.~\ref{tab:compression} describes the influence of compression rate. It demonstrates that HM-ViT can still outperform other methods even under large compression rates. 
\subsection{Qualitative results}
Fig.~\ref{fig:Qualitative} depicts the qualitative visualizations for HM-ViT and No Fusion baselines. In fig.~\ref{fig:Qualitative}a-c, we plot the detection results for camera ego vehicles with no collaborator, camera collaborators, and LiDAR collaborators respectively while for Fig.~\ref{fig:Qualitative}d-f, we plot the results for LiDAR ego vehicles. As shown in these figures, the collaborations with both homogeneous and heterogeneous agents are beneficial for camera ego vehicles and LiDAR ego vehicles with enhanced detection results. In particular, the collaboration between camera ego vehicles and LiDAR collaborators can dramatically enhance the perception performance. 

\section{Conclusion}
In this paper, we present HM-ViT, a hetero-modal vision transformer, for the hetero-modal multi-agent cooperative perception problem which is an important but under-explored research direction. We propose a generic heterogeneous 3D graph attention to jointly reason heterogeneous inter-agent and cross-agent interactions. Our extensive experiments demonstrate the outstanding performance of the proposed method and the great potential of hetero-modal multi-agent collaborations for increasing the scalability and robustness of V2V systems.  We hope our findings and open-source efforts will inspire more research on this new problem. 
{\small
\bibliographystyle{ieee_fullname}
\bibliography{egbib}
}

\end{document}